\documentclass[5p,twocolumn]{elsarticle}

\usepackage{amssymb}

\usepackage{acronym}
\usepackage{multirow}
\usepackage{xurl}

\acrodef{ANN}[ANN]{Artificial Neural Network}
\acrodef{SNN}[SNN]{Spiking Neural Network}
\acrodef{FSM}[FSM]{Finite State Machine}
\acrodef{CSS}[CSS]{Constant Spike Source}
\acrodef{STDP}[STDP]{Spike-Timing-Dependent Plasticity}

\journal{Neural Networks}

\begin{document}

\begin{frontmatter}



\title{Construction of a spike-based memory using neural-like logic gates based on Spiking Neural Networks on SpiNNaker}

\author[inst1]{Alvaro Ayuso-Martinez}
\ead{aayuso@us.es}

\author[inst1]{Daniel Casanueva-Morato}
\author[inst1]{Juan P. Dominguez-Morales}
\author[inst1]{Angel Jimenez-Fernandez}
\author[inst1]{Gabriel Jimenez-Moreno}


\address[inst1]{Robotics and Technology of Computers Lab, Universidad de Sevilla, Av. Reina Mercedes s/n. Escuela Tecnica Superior de Ingenieria Informatica, Sevilla, 41012, Andalucia, Spain}

\begin{abstract}
Neuromorphic engineering concentrates the efforts of a large number of researchers due to its great potential as a field of research, in a search for the exploitation of the advantages of the biological nervous system and the brain as a whole for the design of more efficient and real-time capable applications. For the development of applications as close to biology as possible, Spiking Neural Networks (SNNs) are used, considered biologically-plausible and that form the third generation of Artificial Neural Networks (ANNs). Since some SNN-based applications may need to store data in order to use it later, something that is present both in digital circuits and, in some form, in biology, a spiking memory is needed. This work presents a spiking implementation of a memory, which is one of the most important components in the computer architecture, and which could be essential in the design of a fully spiking computer. In the process of designing this spiking memory, different intermediate components were also implemented and tested. The tests were carried out on the SpiNNaker neuromorphic platform and allow to validate the approach used for the construction of the presented blocks. In addition, this work studies in depth how to build spiking blocks using this approach and includes a comparison between it and those used in other similar works focused on the design of spiking components, which include both spiking logic gates and spiking memory. All implemented blocks and developed tests are available in a public repository.
\end{abstract}




\end{frontmatter}

\section{Introduction}
\label{introduction}

Neuromorphic engineering was presented by Carver Mead in the late 1980s \cite{mead1990neuromorphic}. This concept proposed the development of hardware and software applications based on the fundamental principles of biological nervous systems, one of the most optimal and useful natural mechanisms. Over the years, this concept has given rise to a new and interesting field of research which has experienced a strong evolution in the last two decades \cite{soman2016recent}.

Neuromorphic systems are analog, digital or mixed-signal systems which rely on artificial neurons and spikes to transmit information. In biology, these spikes are large peaks in the membrane potential of neurons that occur when the membrane potential reaches a specific threshold. To mimic this behavior, artificial neurons in neuromorphic systems generate these spikes as asynchronous electric pulses. Thanks to this bio-inspired approach, neuromorphic systems can achieve low power consumption and high real-time capability, which could greatly improve the performance and possibilities of existing systems.

As these neuromorphic systems try to mimic biological nervous systems, it is not only necessary to use bio-inspired neurons, but also bio-inspired architectures. A specific type of biologically-plausible neural networks called \acp{SNN} are commonly used for this purpose. These \acp{SNN} have two basic bio-inspired elements: neurons and synapses. They can be seen as graphs according to mathematical graph theory, where neurons would be nodes and synapses would be edges, which have associated weights (as in other types of neural networks) and delay values. 

There exists two different alternatives to work with \acp{SNN} in order to develop neuromorphic applications: software simulators and hardware platforms. 

Software simulators allow to build and test new applications without implementing a real bio-inspired architecture but rather simulated. Some very popular examples are NEST \cite{Gewaltig:NEST} and Brian \cite{goodman2008brian}.

On the other hand, all neuromorphic hardware platforms are made, in essence, of transistors. Although most of them are fully digital, such as SpiNNaker \cite{furber2014spinnaker}, Loihi \cite{davies2018loihi} or TrueNorth \cite{akopyan2015truenorth}, there exists other platforms, including BrainScaleS \cite{10.3389/fnins.2022.795876}, that use a mixed-signal approach. 

As explained in \cite{furber2016large}, processing in neurons and synapses in the brain uses energy-efficient analog techniques that lack the noise immunity of traditional digital systems and are not fully deterministic, two aspects that are inherent to analog circuits. Since mixed-signal neuromorphic hardware platforms contain analog circuits, they are closer than its digital counterpart to this biological processing in the brain. Using these platforms is a great option to achieve applications that are as bio-inspired as possible and in which great precision in the signal and pure determinism are not needed. For those cases in which those aspects are needed, digital neuromorphic platforms should be considered.

There is a long list of interesting neuromorphic applications that promise great advances in many other different fields. Some of these fields are mentioned in \cite{soman2016recent}, such as image processing \cite{reverter2016neuromorphic} and emotion recognition \cite{diehl2016truehappiness}. Other related works focus on speech recognition \cite{dominguez2018deep, wu2020deep}, sensory fusion \cite{corradi2021gyro, schoepe2020live}, motor control \cite{perez2013neuro, jimenez2012neuro} or bio-inspired locomotion \cite{gutierrez2020neuropod, batres2021biologically}. This evidences the evolution and importance of neuromorphic engineering, whose progress make it possible to think of new and exciting possible future applications, which will require a better understanding of \acp{SNN} and current applications.

As there are no rules on how to build a \ac{SNN} in order to achieve specific behaviors or functionalities, the development of applications that require them may not be a simple task for neuromorphic engineers. To address this, a novel toolkit of functional blocks based on \acp{SNN} was proposed in \cite{ayuso2022spike}, which provides some of the operations usually required. This new toolkit establishes the base for further more complex implementations with similar functionalities than complex digital components. The aforementioned paper also proposed the idea of creating blocks, such as decoders and multiplexers, which would be made from spiking NOT and AND gates, and which would be the base for some higher-level components. In this work, we present as a starting point the design of a spiking decoder, which will be of great importance in the spiking design of one of the most important components of computers: memory. The main objective of this work is to design and implement this spiking memory.

In \cite{sahni2019implementation}, the implementation of neural-like logic gates based on \ac{STDP} \cite{caporale2008spike} is proposed. Moreover, many works which provide implementations of spiking logic gates and circuits based on different paradigms can be found. Spiking Neural P (SN P) systems are used in \cite{song2016design} together with an astrocytes-like control mechanism to build different logic gates, using a unique spiking rule. Other works using these SN P systems are also mentioned there.

On the other hand, no works related to blocks built from spiking logic gates have been found in the literature. Thus, the spike-based blocks presented in this paper are novel. In the case of memory, an implementation based on \ac{STDP} but far from the logic gate approach is presented in \cite{he2019constructing}. It uses Hebb's learning rule \cite{hebb2005organization} to dynamically build connections between two internal layers to achieve the recall ability of biological memory.

In the process of developing the spiking memory, it was previously needed to develop other spiking blocks such as the decoder or the D Latch. Other similar blocks are also presented. The main contributions of this work include the development of the spiking memory and other spiking blocks and the addition of these blocks to the available public repository\footnote{\url{https://github.com/alvayus/sPyBlocks}}, presented in \cite{ayuso2022spike}, together with a set of exhaustive experiments that have been carried out to prove their expected behavior. The SpiNNaker hardware platform is used for running large-scale neural network simulations in real time.

The rest of the paper is structured as follows: Section~\ref{intro_memory} introduces memory design and provides the theoretical basis used to build the spiking memory block; in Section~\ref{technologies}, the software and hardware materials used, as well as some other concepts, are detailed in depth; in Section~\ref{designs}, a list of the proposed new blocks required for the development of the memory block is presented and discussed in depth, showing their associated designs; in Section~\ref{results}, a list of the tests that have been performed is presented with a discussion of their associated results; in Section~\ref{discussion}, a comparison is made between the related works and this work, breaking down the most important differences between the paradigms used. The importance of the proposed blocks in the development of future applications or new components is also discussed in depth; finally, in Section~\ref{conclusions}, the conclusions of the work are presented.

\section{First steps in memory architecture design} 
\label{intro_memory}

Computers need to store large amounts of information, which is divided in instructions and data; programs are made of instructions that use data to perform operations. Thus, memory is a key component within their architecture. In the digital domain, the basic memory unit is the bistable circuit, which is an electric circuit with two stable states, zero and one, that is capable of storing a bit for an indefinite period of time. 

There are two main types of bistable circuits: asynchronous and synchronous circuits, which are commonly called latches and flip-flops, respectively \cite{prasad2021analog}. Moreover, there are different types of latches and flip-flops, whose names usually depend on their inputs and outputs. Commonly, flip-flops are preferred in digital circuits against latches, mainly due to digital systems usually use synchronous subsystems to achieve noise immunity and determinism and, thus, being more reliable than asynchronous systems. The synchronization of the inputs allows to avoid race conditions, in which the time difference between the arrival of inputs could lead to an unwanted state of the circuit.

\begin{figure}[!ht]
\centerline{\includegraphics[height=8cm]{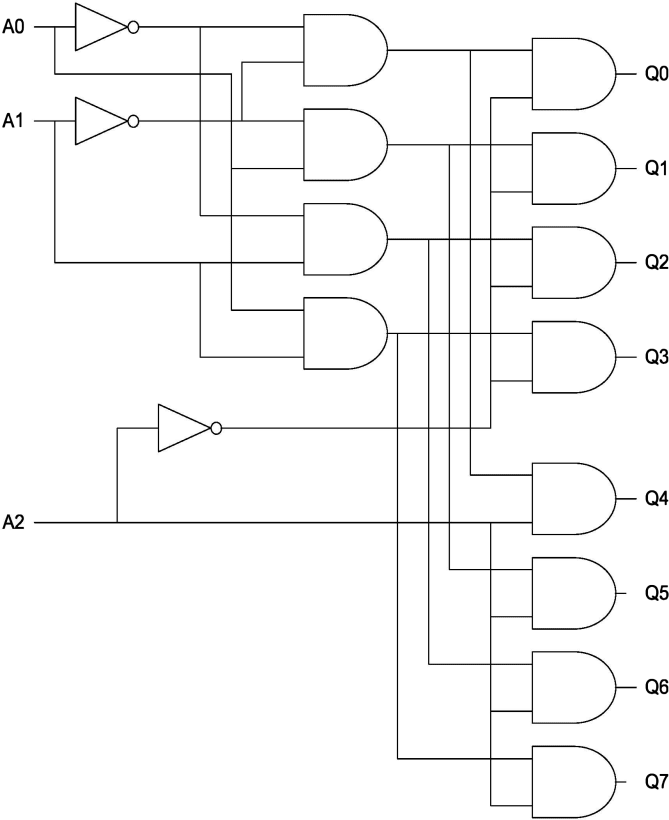}}
\caption{Diagram of the decoder circuit based on NOT and AND gates, extracted from \cite{7019253}. This decoder has 3 inputs and 8 outputs. Note that the AND gates in this figure have a fan-in of 2.}
\label{digital_decoder}
\end{figure}

Registers are arrays of bistable circuits. To have the capability of addressing these registers, a decoder is needed. A decoder is a component with $n$ inputs and $2^n$ outputs, in which the inputs can be seen as the binary representation of a number and the outputs can be seen as a representation of that number in one-hot encoding, which means each output channel represents a unique number. Decoders are built from basic logic gates, usually NOT and AND gates. Its digital circuit is shown in Figure~\ref{digital_decoder}.

\begin{figure}[!ht]
\centerline{\includegraphics[width=\linewidth]{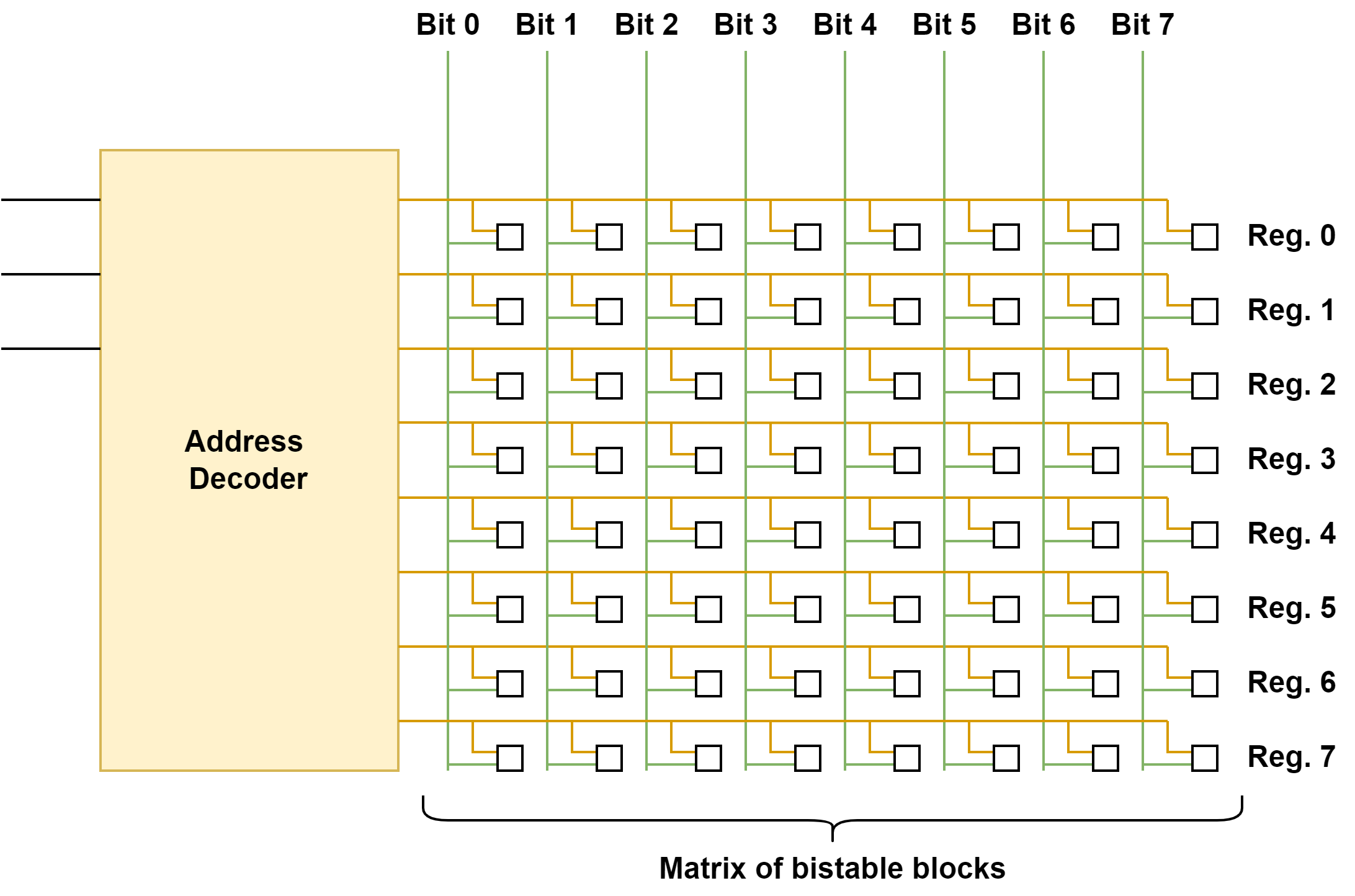}}
\caption{Diagram of the usual memory structure. This example shows a memory of 8 registers, one of each has a width of a byte.}
\label{digital_memory_basic}
\end{figure}

Figure~\ref{digital_memory_basic} shows the usual structure of a memory based on bistable blocks and the decoder mentioned above. Note that this is the basic and early design of a memory architecture in which there are no read and write lines, and no other combinational circuitry to allow these processes. In this case, the decoder outputs are only used as enable signals for the bistable blocks in the memory, represented in Figure~\ref{digital_memory_basic} as small empty squares. The set of these blocks forms a matrix in which the rows represent each register. 

\section{Materials and methods}
\label{technologies}

\subsection{Spiking Neural Networks}

There are currently three different generations of \acp{ANN}. While the second generation is associated to Deep Learning, \acp{SNN} form the basis of the third generation \cite{nandakumar2018building}. All \acp{ANN} share a common structure: they are always built from neurons (nodes) and synapses (connections). This structure is based on the biological nervous system, but has a different level of abstraction in the third generation: while the first two generations only preserve the structure (neurons and synapses without deeply considering its biological aspects), \acp{SNN} are characterized for the use of bio-inspired neuron and synapse models. Thus, the first two generations are based on highly simplified brain dynamics \cite{ghosh2009spiking} and \acp{SNN} are the closest current approach of neural networks to biological functioning \cite{davidson2021comparison}. 

Another important aspect in \acp{SNN} is the way in which the information is transmitted. In the biological nervous system, information is transmitted across synapses in the form of spikes, which are large peaks in the membrane potential of neurons that occur when the membrane potential reaches a certain threshold potential. Spikes are generated in neurons and then propagated to other neurons across synapses. In the artificial approach, which is similar to its biological counterpart, these spikes are represented by asynchronous electric pulses. 

Although \acp{SNN} are more complex than \acp{ANN}, information coded in spikes makes them more energically efficient as they deal with precise timing, having a low computational cost. This precise timing is also related to sparse coding, which means that spike-rate is usually low. This justifies the low power consumption of \acp{SNN} \cite{davidson2021comparison}. Some improvements in the hardware implementation, such as avoiding multiplications, processing spikes using shifts and sums, and only transmitting single bits of information instead of real numbers, allow achieving real-time execution \cite{lobo2020spiking}.

\subsection{Neuromorphic hardware platform}

As mentioned in Section~\ref{introduction}, the SpiNNaker platform has been used to design new functional blocks and test their correct operation. SpiNNaker is a massively-parallel multi-core computing system which was designed to allow modelling very large \acp{SNN} in real time and whose interconnected architecture is inspired by the connectivity characteristics of the mammalian brain \cite{furber2014spinnaker}.

Both SpiNN-3 and SpiNN-5 machines have been used in this work. The main difference between them is the number of chips: while SpiNN-3 consists of 4 chips, SpiNN-5 has 48 chips. Each of the chips is made up of 18 ARM968E-S cores operating at 200 MHz. More details regarding these machines can be found in \cite{rowley2019spinntools}.

Since our designs try to use an optimal amount of resources, all of them can be simulated in both platforms. On the other hand, to perform simulations of blocks that require more inputs, more outputs or a greater number of internal blocks, as for example in the case of the memory, whose capacity depends on the number of internal D latches, a SpiNN-5 machine is used.

\subsection{Software packages}

PyNN \cite{davison2009pynn} is a Python package for the simulator-independent specification of neuronal network models. Currently, PyNN supports NEURON \cite{hines1997neuron}, NEST \cite{Gewaltig:NEST} and Brian \cite{goodman2008brian} as neural network software simulators, and the SpiNNaker \cite{furber2014spinnaker} and BrainScaleS neuromorphic hardware systems. Thanks to this Python package, the whole code can be executed in all supported simulators and hardware platforms. In this work, PyNN 0.9.6 is used.

Another important software package is SpyNNaker \cite{rhodes2018spynnaker}, which is required to work with PyNN and the SpiNNaker hardware platform. In this work, SpyNNaker 6.0.0 is used.

Other software packages that have been used to test the implemented designs are Matplotlib 3.5.1, XlsxWriter 3.0.2, and Numpy 1.22.1.

\subsection{Neuron parameters}

All functional blocks based in \acp{SNN} presented in this work are made to be as independent as possible from the neuron parameters that have been used. Thus, their functionality is based on the behavior of the network as a whole, rather than on individual neurons. In this way, static synapses with different weights and delays are used to achieve specific behaviors, thanks to which it can be ensured that the designed functional blocks will work as intended in any case, disregarding the neuron parameters. Note that, since static synapses are the basis for making these designs, there is no learning involved. 

However, although the proposed designs were made to be as independent as possible from the parameters of the neurons, there are two fundamental details that must be taken into account when selecting these parameters:

Firstly, one input spike must make neurons fire once. This is really important for the designed blocks to work. On the other hand, since only one spike is needed to know the result of the operation performed by the block, if the output response of the neurons contains more than one spike, performance would be diminished since output responses would be longer and the sets of inputs of consecutive operations should be separated in time to avoid the overlap of output responses, which could lead to unexpected behaviors of the functional blocks. Thus, it must be ensured that the expected set of inputs produces exactly one output spike, which will allow avoiding the decrease in the efficiency of the designed blocks, i.e., the decrease in the operating frequency.

\begin{figure}[!ht]
\centerline{\includegraphics[width=\linewidth]{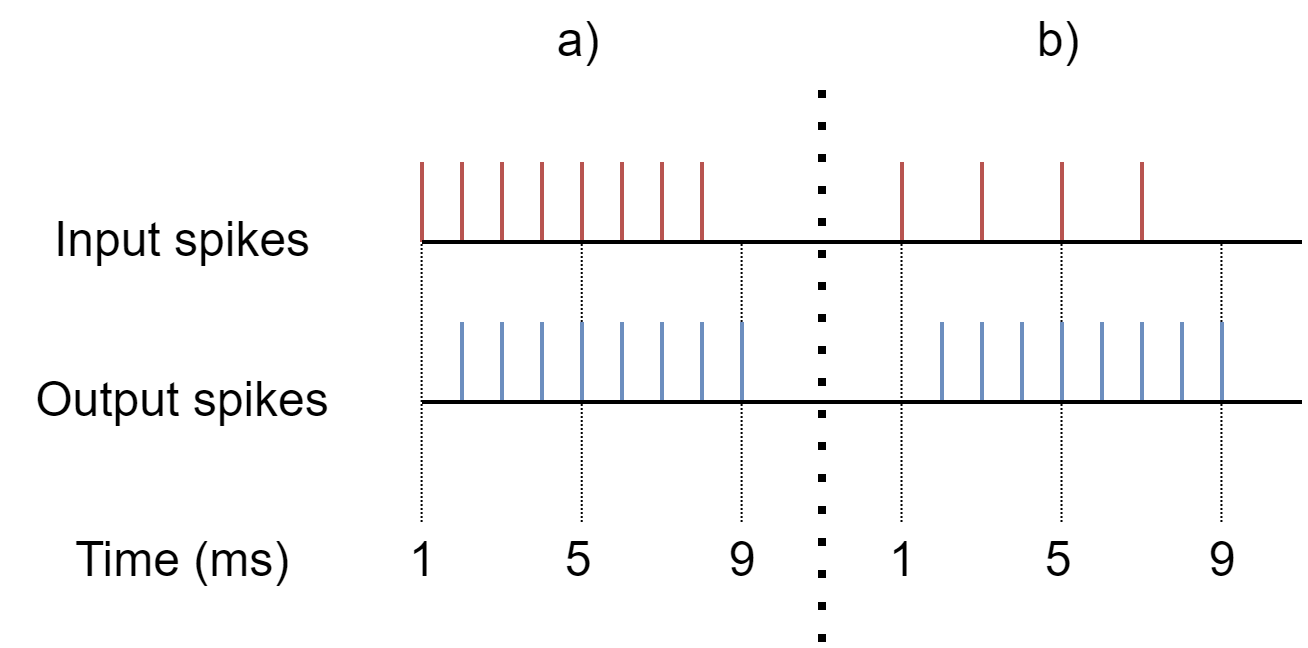}}
\caption{Graphical comparison between shorter and longer output responses. a) Shortest output response (1 output spike per input spike). b) Longer output response (2 output spikes per input spike).}
\label{figure_output_spikes}
\end{figure}

Figure \ref{figure_output_spikes} shows the optimal output response at the left of the dotted line (a), in which a neuron fires one output spike per input spike. Note that there is a latency of 1 ms and there is no overlap between the spikes in the output response. At the right of the dotted line (b), the output response of a neuron that fires two output spikes per input spike is shown. Since firing every millisecond would cause an overlap in the output spikes, input spikes have been separated in time. In this way, these are separated as many milliseconds as output spikes are contained in the output response for a single operation, in this case, 2 ms.

Second, it is necessary to ensure that, after firing, these neurons do not receive any more input spikes until they return to the resting potential, which is necessary to correctly perform the next operation.

In \cite{ayuso2022spike}, the best set of neuron parameters for the SpiNNaker platform was explored. This set fulfills the two previously mentioned requirements and, thus, it has been used in this work. Table~\ref{neuron_parameters} shows and explains these neuron parameters, which might change for other neuromorphic platforms or simulators as the models of the neurons or the meaning of their parameters may vary.

\begin{table}
    \centering
    \caption{Set of neuron parameters used for the implementation of the proposed functional blocks.}
    \label{neuron_parameters}
    \resizebox{\linewidth}{!}{\begin{tabular}{|c|c|c|}
    \hline
    
    \textbf{Neuron parameter} & \textbf{Overview} & \textbf{Value} \\ 
    
    \hline
    $c_m$ & Membrane capacitance & 0.1 nF \\
    
    \hline
    $tau_m$ & Time-constant of the RC circuit & 0.1 ms \\
    
    \hline
    $tau_{refrac}$ & Refractory period & 1.0 ms \\
    
    \hline
    $tau_{syn\_E}$ & Excitatory input current decay time-constant & 0.1 ms \\
    
    \hline
    $tau_{syn\_I}$ & Inhibitory input current decay time-constant & 0.1 ms \\
    
    \hline
    $v_{rest}$ & Resting potential & -65.0 mV \\
    
    \hline
    $v_{reset}$ & Reset potential & -65.0 mV \\
    
    \hline
    $v_{thresh}$ & Threshold potential & -64.91 mV \\
    
    \hline
    \end{tabular}}
\end{table}

\section{Designs} 
\label{designs}

Logic gates contained in a digital component can be replaced by their associated spiking building blocks in order to achieve a higher-level spiking component that performs the same function as the original digital circuit. This is explained in \cite{ayuso2022spike}, where these spiking building blocks, or spiking logic gates, were introduced. Both spiking building blocks and higher-level spiking components are also named functional blocks.

As explained in the previous sections, the main objective of this work is to provide a spiking implementation of a memory block, a process that starts with the development of a spiking decoder. As a counterpart to the decoder, the encoder block, which fulfills its inverse function, is also implemented. Moreover, the circuit of digital decoders is very similar to the circuit of digital multiplexers and demultiplexers. Both components have been implemented in their spiking form due to their implementation being practically immediate from the decoder block and all of them are a good complement to the decoder.

In \cite{ayuso2022spike}, spiking SR Latches were presented as the starting point for storing spiking information. Since biological nervous systems do not seem to use any kind of clock signal, the use of asynchronous circuits, such as these latches, seems the most convenient. In this way, digital bistable blocks in Figure~\ref{digital_memory_basic} should be replaced by new spiking blocks with a similar function, which would be spiking SR Latches enhanced with the capability to store a bit value at a specific time. These new blocks are called spiking D latches. Thus, D latches are also implemented in this work, and therefore the list of implemented blocks is: Decoder, Encoder, Multiplexer, Demultiplexer, D Latch and Memory.

\begin{figure*}[!ht]
\centerline{\includegraphics[width=\linewidth]{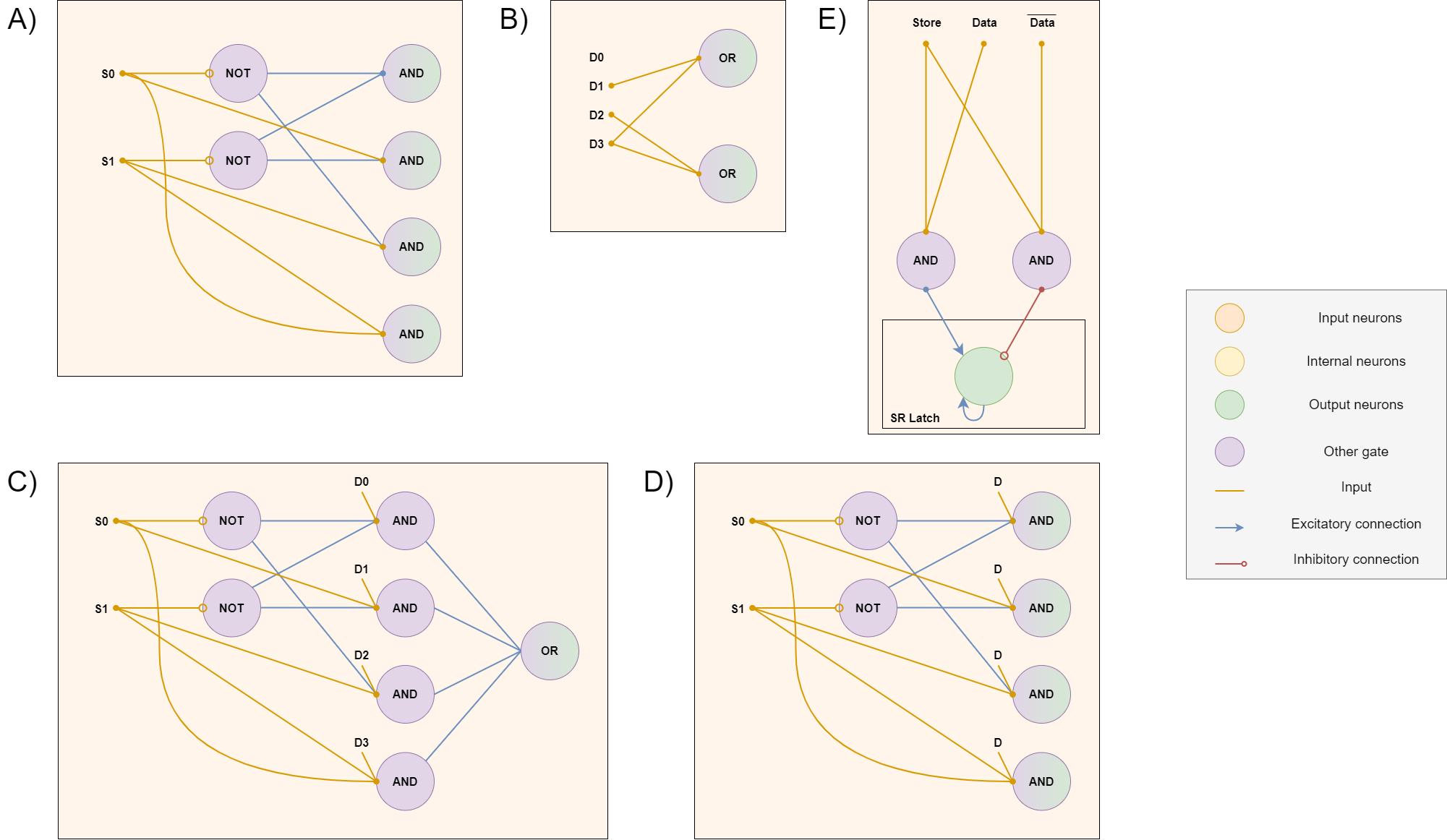}}
\caption{Diagram showing the design of each implemented block. The legend on the right shows the meaning for each color and symbol. \textbf{A)}~Decoder. \textbf{B)} Encoder. \textbf{C)} Multiplexer. \textbf{D)} Demultiplexer. \textbf{E)} D Latch.}
\label{component_designs}
\end{figure*}

The designs of these new blocks can be seen in Figure~\ref{component_designs}, together with a legend of the meaning of each color and symbol used, except for the spiking memory block, whose design is presented in Figure~\ref{spiking_memory}. 

The development process of the spiking implementations of decoder, encoder, multiplexer and demultiplexer is based on the truth table of each associated digital circuit. To calculate the total resources used in each implemented block, the resources used by each block contained in them together with the number of blocks of each type are taken into account.

\subsection{Decoder}
\label{decoder}

As explained in Section~\ref{intro_memory}, a decoder takes its inputs as a binary representation of a number and outputs its one-hot representation in individual channels. In Figure~\ref{digital_decoder}, the circuit of a basic digital decoder built from NOT gates and AND gates with a fan-in of 2 is shown. Replacing the digital NOT and AND gates by spiking NOT and classic or fast AND gates presented in \cite{ayuso2022spike}, a spiking design of the binary decoder can be achieved. This design is shown in the Figure~\ref{component_designs}A.

$S0$ and $S1$ input signals shown in the design are control signals: their activation implies the activation of one output AND, also referred as output channel, which emits a single spike for a single set of input spikes.

The pattern of connections with each of the output AND gates is directly based on the truth table of the decoder. Table~\ref{decoder_truth_table} is the truth table of a decoder with 2 inputs and 4 outputs. For each row, values in the input columns ($S0$ and $S1$) represent the connections with a single AND output gate. While ``1s'' imply connections directly from the input, ``0s'' imply connections from their negated values, i.e., from their associated NOT gates. These connections target the AND gate whose column is ``1''. Thus, e.g., the second row ($\overline{S1}$ and $S0$) connects $S0$ and the NOT gate associated with $S1$ to AND gate 1.

\begin{table}
    \centering
    
    \caption{Truth table of the decoder block design shown in Figure~\ref{component_designs}A. $S1$ and $S0$ are input columns, while AND are output columns.}
    
    \label{decoder_truth_table}
    
    \begin{tabular}{|c|c|c|c|c|c|}
    \hline
    
    {S1} & \textbf{S0} & \textbf{AND 3} & \textbf{AND 2} & \textbf{AND 1} & \textbf{AND 0} \\ 
    
    \hline
    0 & 0 & 0 & 0 & 0 & 1 \\
    \hline
    0 & 1 & 0 & 0 & 1 & 0 \\
    \hline
    1 & 0 & 0 & 1 & 0 & 0 \\
    \hline
    1 & 1 & 1 & 0 & 0 & 0 \\
    
    \hline
    \end{tabular}
\end{table}

When calculating the total number of neurons and synapses used in the development of this spiking implementation, as there are $n$ inputs and $2^n$ outputs in the decoder, there are $n$ NOT and $2^n$ AND gates. One way to count connections from inputs and NOT gates to AND gates is knowing that, since a truth table is used and there are $2^n$ possible combinations with the values of the inputs that arise from alternating the binary values in said inputs, there will be $2^{n - 1}$ connections from each single input and $2^{n-1}$ connections from its associated NOT gate. Note that $2^{n-1} = 2^n / 2$, and that the number of total connections by each input to an AND gate is the number of rows of the truth table, $2^n$.

\begin{itemize}
    \item Nodes
    \begin{itemize}
        \item NOT: $n * 1 = n$
        \item AND (classic): $2^n * 2 = 2^{n+1}$
        \item AND (fast): $2^n * 1 = 2^n$
    \end{itemize}
    
    \item Connections
    \begin{itemize}
        \item Input to NOT: $n * 1 = n$
        \item Input to AND (classic): $n * 2^{n-1} * 2 = n * 2^n$
        \item NOT to AND (classic): $n * 2^{n-1} * 2 = n * 2^n$
        \item Internal AND (classic): $2^n * 1 = 2^n$
        \item Input to AND (fast): $n * 2^{n-1} * 1 = n * 2^{n-1}$
        \item NOT to AND (fast): $n * 2^{n-1} * 1 = n * 2^{n-1}$
    \end{itemize}
\end{itemize}

On the other hand, NOT and fast AND gates need a \ac{CSS} block with provides them with constant spikes. Thus, some more resources are added:

\begin{itemize}
    \item Nodes (CSS): 2
    
    \item Connections (CSS)
    \begin{itemize}
        \item Internal \ac{CSS}: $2$
        \item \ac{CSS} to NOT: $2 * n$
        \item \ac{CSS} to AND (fast): $2 * 2^n = 2^{n+1}$
    \end{itemize}
\end{itemize}

Thus, the number of total neurons depends on the type of AND gate used, and, if the \ac{CSS} resources are included, the result of the calculation is the following:

\begin{itemize}
    \item Total resources (Decoder with classic AND + \ac{CSS})
    \begin{itemize}
        \item Neurons: $2^{n+1} + n + 2$
        \item Synapses: $2^n * (2n + 1) + 3n + 2$
    \end{itemize}
    
    \item Total resources (Decoder with fast AND + \ac{CSS})
    \begin{itemize}
        \item Neurons: $2^n + n + 2$
        \item Synapses: $2^n * (n + 2) + 3n + 2$
    \end{itemize}
\end{itemize}

Where $n$ is the number of inputs of the decoder block. Note that the use of fast AND gates reduces the number of neurons and synapses required for this design.

\subsection{Encoder}

The encoder performs the inverse function of the decoder block, that is, converting from one-hot representation to binary representation. A spiking implementation of this block requires only OR gates, as shown in Figure~\ref{component_designs}B. Notice that, in this figure, there are 4 inputs and 2 outputs, which matches the number of outputs and inputs of the decoder block, respectively. 

In this case, the number of neurons needed can be calculated directly. Since spiking OR gates are a single neuron, the number of neurons is the number of outputs. Being $n$ the number of inputs and $2^n$ the number of outputs in the decoder block, the number of inputs and outputs in the encoder block is $2^n$ and $n$, respectively. As can be seen, the numbers are reverted.

Let $n$ be the number of inputs in the encoder block, the number of outputs in this block could be considered as $\lceil log_2 n \rceil$. Thus, the number of neurons in the encoder block is $\lceil log_2 n \rceil$. The ceiling function has to be used to correctly output the binary representation of the set of inputs, which represents a value in one-hot encoding.

Table~\ref{encoder_truth_table}, which shows the truth table of the encoder block with 4 inputs and 2 outputs, is used to calculate the number of synapses the encoder block contains. Note that the last two columns in the table show ``1s'' when the input is connected to the OR block specified at the top of the column, and, thus, the number of synapses can be calculated as the number of ``1s'' in these columns. This procedure is very similar to the one already explained for the decoder, but, in this case, there is no need to count the zeros because there are no NOT gates. Furthermore, the truth table of the encoder is the truth table of the decoder but reversed, which makes sense since this block performs the inverse function.

\begin{table}
    \centering
    
    \caption{Truth table of the encoder block design shown in Figure~\ref{component_designs}B. $D3$, $D2$, $D1$ and $D0$ are input columns, while OR are output columns. Other unexpected cases will be the combination of those presented here.}
    
    \label{encoder_truth_table}
    
    \begin{tabular}{|c|c|c|c|c|c|}
    \hline
    
    \textbf{D3} & \textbf{D2} & \textbf{D1} & \textbf{D0} & \textbf{OR 1} & \textbf{OR 0} \\ 
    
    \hline
    0 & 0 & 0 & 0 & 0 & 0 \\
    \hline
    0 & 0 & 0 & 1 & 0 & 0 \\
    \hline
    0 & 0 & 1 & 0 & 0 & 1 \\
    \hline
    0 & 1 & 0 & 0 & 1 & 0 \\
    \hline
    1 & 0 & 0 & 0 & 1 & 1 \\
    
    \hline
    \end{tabular}
\end{table}

In a similar way to what was explained in Section~\ref{decoder}, there are $2^n$ possible binary values represented by the output columns of Table~\ref{encoder_truth_table}. However, the number of inputs cannot be assumed to be a power of two, and thus the number of ones is not necessarily $2^{n}/2$ in each of those columns. The formulas to calculate the total number of resources in this block are as follows:

\begin{itemize}
    \item Total resources (Encoder)
    \begin{itemize}
        \item Neurons: $\lceil log_2 n \rceil$
        \item Synapses: $\sum_{i = 2}^{n} ones(bin(i - 1))$
    \end{itemize}
\end{itemize}

Where $n$ is the number of inputs of the block and $ones(bin(i-1))$ is the number of ones of the binary representation of $i-1$, where $i$ represents the rows of the table. Note that this equation ignores the first two rows because they do not contribute anything to the output. In Table~\ref{encoder_truth_table}, second row has one input but no outputs because it is associated to the non-operation output channel of the decoder block. In the encoder design shown in Figure~\ref{component_designs}B, D0 represents this first non-operation input, and is not connected to achieve the desired behavior. 

\subsection{Multiplexer}

A multiplexer is a component that allows an input signal to be selected using control signals, which is why it is commonly referred to as a data selector. It is very similar to decoder, but has two differences: firstly, while a decoder only has channel selection inputs, a multiplexer also has data inputs, which are directly connected to the output AND gates. Secondly, since multiplexers must have only one output channel, there is an OR gate that sums all the signals emitted by the AND gates in a single channel. 

In the spiking multiplexer design, shown in Figure~\ref{component_designs}C, both differences can be seen graphically. Table~\ref{multiplexer_truth_table}, which is very similar to Table~\ref{decoder_truth_table}, represents the truth table of this design. 

\begin{table}
    \centering
    
    \caption{Truth table of the multiplexer block design shown in Figure~\ref{component_designs}C. $S1$ and $S0$ are input columns, while OR is output column.}
    
    \label{multiplexer_truth_table}
    
    \begin{tabular}{|c|c|c|}
    \hline
    
    {S1} & \textbf{S0} & \textbf{OR} \\ 
    
    \hline
    0 & 0 & D0 \\
    \hline
    0 & 1 & D1 \\
    \hline
    1 & 0 & D2 \\
    \hline
    1 & 1 & D3 \\
    
    \hline
    \end{tabular}
\end{table}

Since multiplexers are very similar to decoders, calculating the number of resources required will be much easier as there is an starting point, which is the number of neurons and synapses required to build the spiking decoder mentioned above. Therefore, it is only necessary to calculate the number of added neurons and synapses, which are the following:

\begin{itemize}
    \item Added neurons: 1
    
    \item Added synapses
    \begin{itemize}
        \item Data inputs to AND (classic): $2^n * 2$
        \item Data inputs to AND (fast): $2^n$
        \item AND to OR: $2^n$
    \end{itemize}
\end{itemize}

Thus, the total resources required are as follows:

\begin{itemize}
    \item Total resources (Multiplexer with classic AND + \ac{CSS})
    \begin{itemize}
        \item Neurons: $2^{n+1} + n + 3$
        \item Synapses: $2^n * (2n + 4) + 3n + 2$
    \end{itemize}
    
    \item Total resources (Multiplexer with fast AND + \ac{CSS})
    \begin{itemize}
        \item Neurons: $2^n + n + 3$
        \item Synapses: $2^n * (n + 4) + 3n + 2$
    \end{itemize}
\end{itemize}

\subsection{Demultiplexer}

To perform the inverse function of a multiplexer, the resulting circuit of a demultiplexer is also very similar to that of a decoder. There is one difference: as in the case of multiplexer, there are data inputs directly connected to the output AND gates too, but, in this case, these inputs are actually the same, which is supposed to be the output of a multiplexer. Note that since there must be multiple output channels in a demultiplexer, the OR gate introduced in the multiplexer design is no longer needed. The spiking demultiplexer design is shown in Figure~\ref{component_designs}D.

The truth table of this design is shown in 
Table~\ref{demultiplexer_truth_table}. Notice that this truth table is very similar to the decoder's (Table~\ref{decoder_truth_table}), but the output values of the AND gates now depend on the value of the data input.

\begin{table}
    \centering
    
    \caption{Truth table of the demultiplexer block design shown in Figure~\ref{component_designs}D. $S1$ and $S0$ are input columns, while AND are output columns.}
    
    \label{demultiplexer_truth_table}
    
    \begin{tabular}{|c|c|c|c|c|c|}
    \hline
    
    {S1} & \textbf{S0} & \textbf{AND 3} & \textbf{AND 2} & \textbf{AND 1} & \textbf{AND 0} \\ 
    
    \hline
    0 & 0 & 0 & 0 & 0 & D \\
    \hline
    0 & 1 & 0 & 0 & D & 0 \\
    \hline
    1 & 0 & 0 & D & 0 & 0 \\
    \hline
    1 & 1 & D & 0 & 0 & 0 \\
    
    \hline
    \end{tabular}
\end{table}

As commented before, although the demultiplexer must perform the inverse operation of the multiplexer, this spiking design is very similar to that of the multiplexer. In this way, both designs have the same number of input connections, despite the previously-mentioned difference. 

While there are no added neurons regarding the spiking design of the decoder, since the OR gate introduced in multiplexer is not introduced here, added synapses between data input and AND gates must be taken into account:

\begin{itemize}
    \item Added synapses
    \begin{itemize}
        \item Data inputs to AND (classic): $2^n * 2 = 2^{n+1}$
        \item Data inputs to AND (fast): $2^n$
    \end{itemize}
\end{itemize}

The number of total resources required for building this spiking demultiplexer are calculated from the number of total resources of the decoder block, summing these new synapses. Thus, the result of the calculation is as follows:

\begin{itemize}
    \item Total resources (Demultiplexer with classic AND + \ac{CSS})
    \begin{itemize}
        \item Neurons: $2^{n+1} + n + 2$
        \item Synapses: $2^n * (2n + 3) + 3n + 2$
    \end{itemize}
    
    \item Total resources (Demultiplexer with fast AND + \ac{CSS})
    \begin{itemize}
        \item Neurons: $2^n + n + 2$
        \item Synapses: $2^n * (n + 3) + 3n + 2$
    \end{itemize}
\end{itemize}

\subsection{D Latch}

As explained in this section, D Latch is required for the design of a memory block and can be built from an SR Latch. While SR latches need a spike through their set connections to set the latch and through their reset connections to reset it, D latches need a spike through their store signal connection to store the value (1 if there is a spike, 0 if not) received through their data connection.

This new design of a latch is required to build a memory block since dealing with register addressing is needed, which implies a control signal. That control signal will be used to perform write operations in D latches thanks to the store signal connection.

The spiking design of a D Latch is shown in Figure~\ref{component_designs}E. Note that there are not only store and data connections, but there is also a connection called $\overline{data}$. This connection is expected to receive the negated value of the data, using an external NOT gate. In the lower part of this design there is a rectangle delimiting a neuron and its associated connections, which corresponds to the design of an SR Latch. 

The number of neurons and synapses needed are then calculated:

\begin{itemize}
    \item Neurons
    \begin{itemize}
        \item AND (classic): $2 * 2 = 4$
        \item AND (fast): $2 * 1 = 2$
        \item SR Latch: $1 * 1 = 1$
    \end{itemize}
    
    \item Synapses
    \begin{itemize}
        \item Store to AND (classic): $2 * 2 = 4$
        \item Data to AND (classic): $1 * 2 = 2$
        \item $\overline{Data}$ to AND (classic): $1 * 2 = 2$
        \item Internal AND (classic): $2 * 1 = 2$
        \item Store to AND (fast): $2 * 1 = 2$
        \item Data to AND (fast): $1 * 1 = 1$
        \item $\overline{Data}$ to AND (fast): $1 * 1 = 1$
        \item AND to SR Latch (set): $1 * 1 = 1$
        \item AND to SR Latch (reset): $1 * 1 = 11$
        \item Internal SR Latch: $1 * 1 = 1$
    \end{itemize}
\end{itemize}

Note that the previous calculation has statements in the form \textit{number of blocks * number of neurons in the block} or \textit{number of blocks * number of connections in the block}. A \ac{CSS} block should also be introduced to allow the correct operation of fast AND gates. To facilitate the calculation of resources used for the memory design, CSS neurons and synapses are not included in the total resources of a single D Latch, which are the following:

\begin{itemize}
    \item Total resources (D Latch with classic AND)
    \begin{itemize}
        \item Neurons: $5$
        \item Synapses: $13$
    \end{itemize}
    
    \item Total resources (D Latch with fast AND. CSS block not included)
    \begin{itemize}
        \item Neurons: $3$
        \item Synapses: $7$
    \end{itemize}
\end{itemize}

The amount of total resources needed to build a D Latch is higher than an SR Latch since it adds AND gates to achieve the desired behavior.

\subsection{Memory}
\label{memory_design_section}

\begin{figure*}[!ht]
\centerline{\includegraphics[height=11cm]{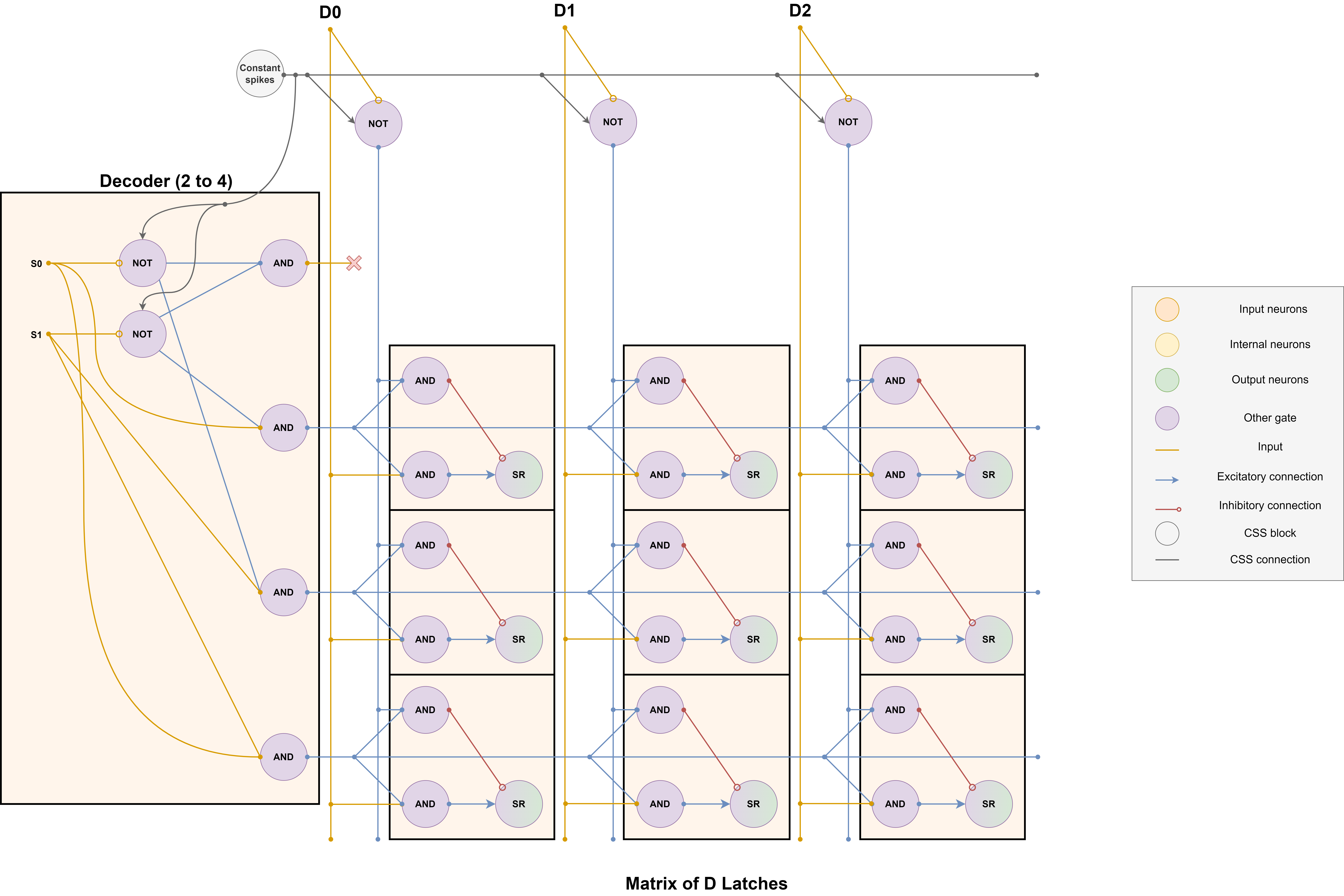}}
\caption{Example design of a spiking memory block with 3 words (rows) of 3 bits (columns) using classic AND gates.}
\label{spiking_memory}
\end{figure*}

The memory block was introduced in Section~\ref{intro_memory}. While Figure~\ref{digital_memory_basic} shows the basic digital circuit for a memory block, replacing its digital components with the spiking ones presented in this section is enough to reach the associated spiking design, which is shown in Figure~\ref{spiking_memory}.

This memory block is mainly made up of one decoder and a matrix of D latches. The first output channel of the decoder corresponds to the non-operation channel, so it will be constantly firing spikes when no input spikes are received. Using these output spikes would mean that the associated register (an array of D latches, a row in the matrix of D latches) would be constantly performing writing operations. Because of this, this channel is not used and there is no register associated with it.

Notice that in Figure~\ref{spiking_memory} there are rectangles delimiting both the decoder and each D Latch. Moreover, there are additional NOT gates which have the function of negating each data bit on the top of the design. The negate output is used by all D latches in the same column of the matrix of D latches. 

In Section~\ref{decoder}, the number of total resources required for building a spiking decoder together with the needed CSS block is shown. These resources are taken as the starting point for the calculation of the total resources of this memory block. Let $r$ be the number of registers (rows) and $c$ their width, i.e., the number of bits of each register (columns), and the total number of D latches $r * c$. As CSS block is already included, it is only needed to sum the following resources:

\begin{itemize}
    \item Added neurons
    \begin{itemize}
        \item NOT: $c * 1 = c$
        \item D latches (classic AND): $r * c * 5$
        \item D latches (fast AND): $r * c * 3$
    \end{itemize}
    
    \item Added synapses
    \begin{itemize}
        \item Data to NOT: $c * 1$
        \item CSS to NOT: $2 * c$
        \item D latches (classic AND): $r * c * 13$
        \item D latches (fast AND): $r * c * 7$
        \item CSS to AND (fast AND): $r * c * 2 * 2$
    \end{itemize}
\end{itemize}

Note that the number of registers is equal to the number of outputs of the decoder minus one, since the first output of the decoder has no register associated with it. Thus, $r = 2^n - 1$. As it is more common to work with the number of registers and their width, $n$ can be isolated from the formula. Then, the following expression defines the number of inputs of the decoder: $n = \lceil log_2 (r + 1) \rceil $. The use of the ceiling function is required to address all decoder output channels. 

Replacing $n$ in the last two formulas presented in Section~\ref{decoder}, the total amount of  resources needed to build the decoder and CSS blocks based on the number of registers, $r$ (not in the number of inputs), can be calculated. The resulting formulas, simplified, are as follows:

\begin{itemize}
    \item Total resources (Decoder with classic AND + \ac{CSS})
    \begin{itemize}
        \item Neurons: $2r + \lceil log_2 (r+1) \rceil + 4$
        \item Synapses: $r + (2r + 5) \lceil log_2 (r+1) \rceil + 3$
    \end{itemize}
    
    \item Total resources (Decoder with fast AND + \ac{CSS})
    \begin{itemize}
        \item Neurons: $r + \lceil log_2 (r+1) \rceil + 3$
        \item Synapses: $2r + (r + 4) \lceil log_2 (r+1) \rceil + 4$
    \end{itemize}
\end{itemize}

Adding the previously calculated resources to these formulas, the final formulas for calculating the total resources of the memory block can be obtained:

\begin{itemize}
    \item Total resources (Memory with classic AND + \ac{CSS})
    \begin{itemize}
        \item Neurons: $2r + c + 5rc + \lceil log_2 (r+1) \rceil + 4$
        \item Synapses: $r + 3c + 13rc + (2r + 5) \lceil log_2 (r+1) \rceil + 3$
    \end{itemize}
    
    \item Total resources (Memory with fast AND + \ac{CSS})
    \begin{itemize}
        \item Neurons: $r + c + 3rc + \lceil log_2 (r+1) \rceil + 3$
        \item Synapses: $2r + 3c + 11rc + (r + 4) \lceil log_2 (r+1) \rceil + 4$
    \end{itemize}
\end{itemize}

The current memory implementation does not use read signals to allow the internal values of the D latches to be read, which is a detail that could be interesting to implement in the near future to allow the memory block to be connected with other implemented spiking functional blocks or design higher level spiking functional blocks. For now, the internal value of D latches should be read immediately with a direct connection from their output.

\subsection{Resources comparison}

Table~\ref{resource_comparison_n} groups all the formulas to calculate the resources of the spiking functional blocks presented in this work based on the number of inputs $(n)$, while Table~\ref{resource_comparison_m} groups the formulas to calculate the resources of said blocks depending on the number of outputs $(m)$. Both tables are presented to facilitate the calculation of the resources used in any possible case. 

In the case of the memory block, the formula depends on $r$, which is the number of registers (rows of the matrix of D latches). $m$ and $r$ are defined by the following expressions: 

\begin{itemize}
    \item $m = 2^{n}$ for the decoder, multiplexer and demultiplexer
    \item $m = log_2 n$ for the encoder
    \item $r = 2^{n} - 1$ for the memory
\end{itemize}

Note that the number of resources of the memory block also depends on $c$, which is the number of bits (columns of the matrix of D latches).

All formulas in Table~\ref{resource_comparison_n} have been explained in this section, except the one associated with the memory block. In the same way, only the formulas associated with this memory block in Table~\ref{resource_comparison_m} have been explained. The rest of the formulas are obtained by substituting the variables according to the previous formulas and are presented for the sake of simplicity. These formulas assume that all inputs are properly connected, as shown in the corresponding designs.

\begin{table*}[ht]
    \centering
    
    \caption{Formulas to obtain the resources required for the design of the different blocks shown in Figure~\ref{component_designs} as a function of the number of inputs $(n)$.}
    
    \label{resource_comparison_n}
    
    \begin{tabular}{|c|c|c|c|}
    \hline
    
    Block & AND type & Total neurons & Total synapses \\ 
    
    \hline
    \multirow{2}{*}{Decoder + CSS} & Classic & $2^{n+1} + n + 2$ & $2^n * (2n + 1) + 3n + 2$ \\
    \cline{2-4}
    & Fast & $2^n + n + 2$ & $2^n * (n + 2) + 3n + 2$ \\
    
    \hline
    Encoder & - & $\lceil log_2 n \rceil$ & $\sum_{i = 2}^{n} ones(bin(i - 1))$ \\
    
    \hline
    \multirow{2}{*}{Multiplexer + CSS} & Classic & $2^{n+1} + n + 3$ & $2^n * (2n + 4) + 3n + 2$ \\
    \cline{2-4}
    & Fast & $2^n + n + 3$ & $2^n * (n + 4) + 3n + 2$ \\
    
    \hline
    \multirow{2}{*}{Demultiplexer + CSS} & Classic & $2^{n+1} + n + 2$ & $2^n * (2n + 3) + 3n + 2$ \\
    \cline{2-4}
    & Fast & $2^n + n + 2$ & $2^n * (n + 3) + 3n + 2$ \\
    
    \hline
    \multirow{2}{*}{D Latch} & Classic & $5$ & $13$ \\
    \cline{2-4}
    & Fast & $3$ & $7$ \\
    
    \hline
    \multirow{2}{*}{Memory + CSS} & Classic & $2^n (5c + 2) + n - 4c + 2$ & $2^n * (2n + 13c + 1) + 3n - 10c + 2$ \\
    \cline{2-4}
    & Fast & $2^n (3c + 1) + n - 2c + 2$ & $2^n (n + 11c + 2) + 3n - 8c + 2$ \\
    
    \hline
    \end{tabular}
\end{table*}

\begin{table*}[ht]
    \centering
    
    \caption{Formulas to obtain the resources required for the design of the different blocks shown in Figure~\ref{component_designs} as a function of the number of outputs $(m)$ or registers $(r)$.}
    
    \label{resource_comparison_m}
    
    \begin{tabular}{|c|c|c|c|}
    \hline
    
    Block & AND type & Total neurons & Total synapses \\ 
    
    \hline
    \multirow{2}{*}{Decoder + CSS} & Classic & $2m + \lceil log_2 m \rceil + 2$ & $m + (2m + 3) \lceil log_2 m \rceil + 2$ \\
    \cline{2-4}
    & Fast & $m + \lceil log_2 m \rceil + 2$ & $2m + (m + 3) \lceil log_2 m \rceil + 2$ \\
    
    \hline
    Encoder & - & m & - \\
    
    \hline
    \multirow{2}{*}{Multiplexer + CSS} & Classic & $2m + \lceil log_2 m \rceil + 3$ & $4m + (2m + 3) \lceil log_2 m \rceil + 2$ \\
    \cline{2-4}
    & Fast & $m + \lceil log_2 m \rceil + 3$ & $4m + (m + 3) \lceil log_2 m \rceil + 2$ \\
    
    \hline
    \multirow{2}{*}{Demultiplexer + CSS} & Classic & $2m + \lceil log_2 m \rceil + 2$ & $3m + (2m + 3) \lceil log_2 m \rceil + 2$ \\
    \cline{2-4}
    & Fast & $m + \lceil log_2 m \rceil + 2$ & $3m + (m + 3) \lceil log_2 m \rceil + 2$ \\
    
    \hline
    \multirow{2}{*}{D Latch} & Classic & $5$ & $13$ \\
    \cline{2-4}
    & Fast & $3$ & $7$ \\
    
    \hline
    \multirow{2}{*}{Memory + CSS} & Classic & $2r + c + 5rc + \lceil log_2 (r+1) \rceil + 4$ & $r + 3c + 13rc + (2r + 5) \lceil log_2 (r+1) \rceil + 3$ \\
    \cline{2-4}
    & Fast & $r + c + 3rc + \lceil log_2 (r+1) \rceil + 3$ & $2r + 3c + 11rc + (r + 4) \lceil log_2 (r+1) \rceil + 4$ \\
    
    \hline
    \end{tabular}
\end{table*}

Moreover, Table~\ref{component_delays} shows the delays for each of the presented blocks, depending on the type of the AND block used. These delay values include all delays from the input connection to the output neuron. A connection with a delay of 1 ms has been taken as the standard connection.

\begin{table}[ht]
    \centering
    
    \caption{Delays for each of the presented blocks.}
    
    \label{component_delays}
    
    \begin{tabular}{|c|c|c|}
    \hline
    
    Block & AND type & Delay (ms) \\
    
    \hline
    \multirow{2}{*}{Decoder} & Classic & 3 \\
    \cline{2-3}
    & Fast & 2 \\
    
    \hline
    Encoder & - & 1 \\
    
    \hline
    \multirow{2}{*}{Multiplexer} & Classic & 4 \\
    \cline{2-3}
    & Fast & 3 \\
    
    \hline
    \multirow{2}{*}{Demultiplexer} & Classic & 3 \\
    \cline{2-3}
    & Fast & 2 \\
    
    \hline
    \multirow{2}{*}{D Latch} & Classic & 3 \\
    \cline{2-3}
    & Fast & 2 \\
    
    \hline
    \multirow{2}{*}{Memory} & Classic & 6 \\
    \cline{2-3}
    & Fast & 4 \\
    
    \hline
    \end{tabular}
\end{table}
\section{Results}
\label{results}

In this section, the implementation and correct operation of the components whose designs were presented in Section~\ref{designs} are tested. For this, the following list of experiments was proposed, which includes tests of different complexities: combined components tests (Decoder - Encoder and Multiplexor - Demultiplexor), simple D Latch tests and complex memory tests. Combined component tests prove the correct operation of all the components involved. Therefore, for these cases, no individual tests are presented in order to avoid repeating similar results.

All these tests also serve to demonstrate that the number of total resources used coincides with those calculated theoretically using the formulas presented in Section~\ref{designs}. The practical calculation of these resources, which was done by counting one by one each of the resources added in the simulation phase, is independent of the theoretical calculation.

\subsection{Decoder - Encoder}

To test the correct operation of the spiking decoder and encoder designs, a short test was developed in which a decoder with continuously increasing binary values at its input is connected to an encoder. The encoder outputs were expected to represent the binary values used as input to the decoder, since, as explained, the encoder performs the inverse function of the decoder.

\begin{figure}[!ht]
\centerline{\includegraphics[width=\linewidth]{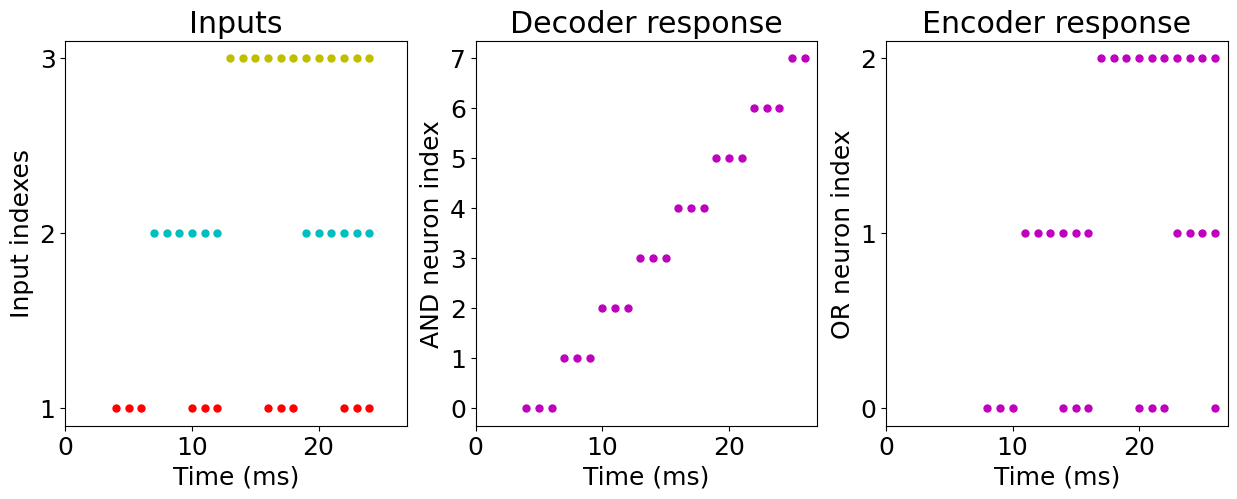}}
\caption{Graphs showing the results of the combined decoder/encoder test.}
\label{decoder_encoder_test}
\end{figure}

Figure~\ref{decoder_encoder_test} shows the results of this test. As expected, increasing the binary input values allows ascending channel selection at the decoder. Also, the encoder outputs are almost the same as the decoder inputs, with only one difference: they are delayed in time. This effect also occurs in the decoder, but with a different delay value. However, the delay value is the same for each of their output channels, for both encoder and decoder. As with digital circuits, these delays are inherent in the spiking designs, with the total output delay being the sum of the delays of all existing connections in the output path.

\subsection{Multiplexer - Demultiplexer}

\begin{figure}[!ht]
\centerline{\includegraphics[width=\linewidth]{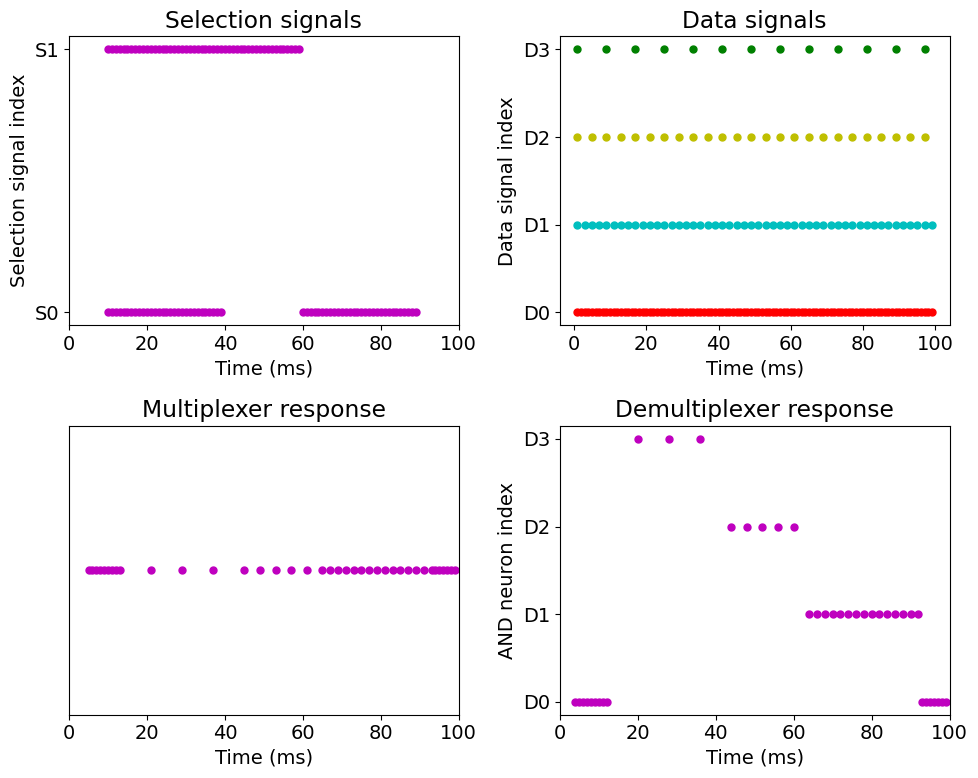}}
\caption{Graphs showing the results of the combined multiplexer/demultiplexer test.}
\label{multiplexer_demultiplexer_test}
\end{figure}

The results of this test are shown in Figure~\ref{multiplexer_demultiplexer_test}. This combined test uses different binary values as control signals that are not related but random. These values are plotted in the upper left subplot, while the data inputs are plotted in the upper right subplot. Of those data inputs, D0 corresponds to the signal that is firing spikes each millisecond, at the bottom of the graph. Successively, D1 has half of the frequency of D0, and so on.

The lower left subplot shows the response of the multiplexer block. Using the signals shown in the graph immediately above, the output signal of the multiplexer is made up of chunks of the data signals. Note that between $t=0$ and $t=10$ there are no spikes in selection inputs, and, as consequence, the multiplexer response is a chunk of D0. After $t=10$, both control signals are active and the multiplexer response is a chunk of D3, which is the signal of the channel associated to value 3 (11 in binary). Other changes in the control signals occur at $t=40$, $t=60$ and $t=90$, which are also reflected as a change in the output of the multiplexer. Notice that these responses are delayed in time since the multiplexer design has an inherent delay. There is only one output signal since the multiplexer has only one output channel.

The lower right subplot shows how the demultiplexer is able to separate each chunk in its original data channel. This is possible thanks to the use of control signals, which indicate which channel the input signal (in this case, the output signal of the multiplexer) should be sent through. Note that these control signals are the same ones used in the multiplexer so that the outputs of the demultiplexer are the same as the data inputs of the multiplexer.

While the demultiplexer has the same latency as the decoder since its output path has the same number of forward connections, the multiplexer adds the delay value of the connection between the AND gate and the output OR gate, being the output path longer in this case.

\subsection{D Latch}

\begin{figure}[!ht]
\centerline{\includegraphics[width=\linewidth]{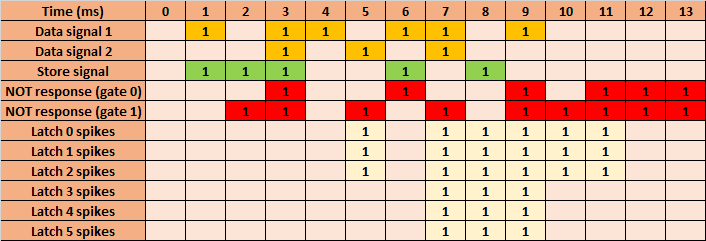}}
\caption{Trace of the D Latch test.}
\label{d_latch_test}
\end{figure}

This test verifies the correct operation of multiple D Latches, which have to be set or reset independently. Figure~\ref{d_latch_test} shows its results in a trace, in which the information can be visualized better than in a graph. In each cell of each row of the trace a ``1'' is displayed when a spike is fired from the corresponding block in the millisecond indicated by the top row. In this case, classic AND gates and external NOT gates were used.

In this experiment, two data bit signals are used, through which spikes are fired with no specific pattern to prove all possible combinations of binary values. Their associated negated values appear as NOT gate responses in the trace, colored in red and delayed by 1 ms from the original time. In the same way, there are spikes released at different timesteps through the store signal. Data signal 1 is associated with the first 3 latches (0, 1 and 2), which are set or reset simultaneously. The same happens with data signal 2, which is associated with latches 3, 4 and 5. Note that there is a delay between the spikes in the data signals and the spikes in the associated latches, which is the delay of the output path. This delay value depends on the AND type used and includes the delay value of the external NOT gate, which, in this case, is equal to 4~ms. Notice that the NOT gate delay is 1~ms, thus the total delay is as expected based on Table~\ref{component_delays}.

The first time that latches 0, 1 and 2 are set to 1 is associated with the first spike of the store signal. This spike at $t=1$ sets latches 0, 1 and 2 at $t=5$. The spike at $t=2$ resets them. In $t=3$, the order to set all the latches is given, which occurs at $t=7$. In $t=4$ and $t=5$ there are no spikes in the store signal, so data spikes are ignored. The rest of the cases are very similar to these. Note that, in $t=8$, the value ``0'' has to be stored. Consequently, in $t=12$ all the latches are reset.

\subsection{Memory}

\begin{figure}[!ht]
\centerline{\includegraphics[width=\linewidth]{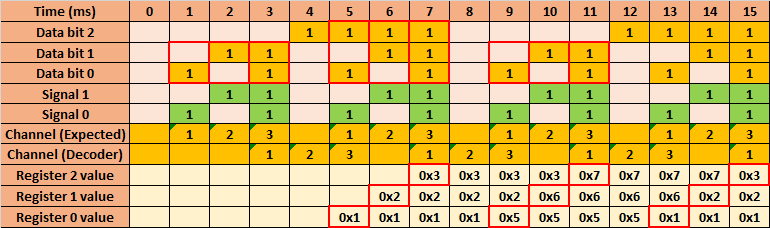}}
\caption{Trace of the memory test.}
\label{memory_test}
\end{figure}

The experiment conducted to test the memory was carried out on the block presented in Figure~\ref{spiking_memory} but using fast AND gates, which means that all AND gates are also connected to the \ac{CSS} block. This test consisted on performing an ascending count through the ``activation'' or ``deactivation'' (existence or not of spikes) of the input signals, and its results are shown in Figure~\ref{memory_test}. The signals colored in green and named ``Signal'' correspond to the decoder's control signals, which indicate the register in which the input must be stored at the moment in which said selection occurs. 

The two rows labeled ``Channel'' correspond to the calculation of the selected channel. In these rows, empty cells correspond to the value 0, which has been ignored to show only the channels of interest. Channel (Expected) shows the selected channel based on the control signal values, while Channel (Decoder) empirically shows the decoder channel that has been activated. Note that there is a delay between the expected channel and the channel actually selected, which is the decoder delay. Thus, the delay between the empirical activation of the decoder output channel and the writing of the register is the latch delay.

In Figure~\ref{memory_test}, each red rectangle on a vertical set of data bits encompasses the binary representation of the count number and is associated with a red rectangle in the register part, at the bottom of the trace, with a delay of 4 ms, which is the delay of this block using fast AND gates. Notice that rectangles in the register part expresses the count number in hexadecimal format. It is necessary to remember that each register has 3 bits, coinciding with the number of input data bits.

At $t=1$, $t=2$ and $t=3$, the values to be written into the register associated with the output channels 1, 2 and 3 of the decoder are 0x01, 0x02 and 0x03 respectively, which are stored at $t=5$, $t=6$ and $t=7$ because of the decoder's delay. At $t=4$ no control signals are provided, and then, at $t=8$, no store operation is performed. This is because the output channel 0 of the decoder has been selected, which does not have any associated register in the memory. In this experiment, the ascending count received at the input data bits reaches it maximum value at $t=7$, and is reset to 0 in the next timestep. Thus, the trace is repeated in a loop until the end of the simulation is reached.
\section{Discussion}
\label{discussion}


In Section~\ref{introduction}, some works are cited in which spiking logic gates were implemented using different approaches. In \cite{sahni2019implementation}, it is proposed the use of STDP. This paradigm needs to use training techniques to define the weights of the neural network synapses, something that does not happen with the static synapses used in this work and in \cite{ayuso2022spike}, as they have fixed weights. Moreover, although STDP could decrease the total number of resources used in the designs presented in \cite{ayuso2022spike}, it would increase the computational cost and, probably, decrease the performance of the designed blocks.

On the other hand, \cite{song2016design} presents SN P systems with an astrocyte-like control mechanism to build neural-like logic gates. This approach is more complex than the \acp{SNN} used in \cite{ayuso2022spike}, since the resulting systems are larger and it additionally introduces a new type of cell: astrocytes. This is easy to see when comparing OR implementations using both approaches, since OR blocks used in this work (presented in \cite{ayuso2022spike}) consists of only a single neuron. Moreover, the behavior of the OR implementation in \cite{song2016design} needs several timesteps to perform the OR operation, which means it has a lower frequency of operation.

SN P systems use spiking and forgetting rules to define the behavior of neurons in the system. In this way, such behavior deviates from the biological models of neurons used in \acp{SNN}, implying that SN P systems are a bio-inspired approach not as bio-plausible as \acp{SNN}, although they can be a very useful alternative.

In addition to everything mentioned above, \cite{song2016design} proposes implementations in a theoretical way. In \cite{ayuso2022spike}, spiking logic gates are not only implemented practically, but are also implemented in hardware, which means going beyond software simulators.

In \cite{he2019constructing} an associative memory based on \acp{SNN} is proposed. This type of memory is characterized by the high correlation between the sources of information, being able to recover all the information only by having a part of it. This is very interesting since it could reduce the computational cost of some certain powerful and complex applications, especially in tasks where it is necessary to find where the information is contained in memory. However, as it dynamically generates new connections by training to define the behavior of the network and uses the STDP paradigm to modify their weights, its computational cost should be higher, as explained before. Its performance, but also the number of resources required, should also be lower than the memory block presented in this work.

Other implementations of associative memories based on \acp{SNN} are also proposed in \cite{casanueva2022spike}, whose strong and weak points are similar to those mentioned above. In addition, these implementations present some limitations in the storage operations which makes them not totally viable for the development of practical applications.


This work focuses on designing new functional blocks based on the building blocks presented in \cite{ayuso2022spike}. These new higher-level blocks provide more complex functionalities as said building blocks and can also be used to build higher-level blocks, gradually achieving more specific functionalities.

In this work, the implementation of spiking D latches was proposed, which allow absolute control over memory operations thanks to the fact that they allow to asynchronously indicate the moment in which the store operation must be carried out. This is very useful in the design of \acp{FSM}, which are made up of combinational logic for transitioning between states and these latches to store their current state. \acp{FSM} can be used to design any sequence of actions, which means that it can be used to build any hardware system with specific functionality. It should be noted that once \acp{FSM} were available, practically any circuit could be implemented. In this way, the possibilities of blocks that can be implemented would be infinite, as in the case of digital circuits.

The implementation of \acp{FSM} is especially useful in the field of embedded systems, where applications such as IoT, elevators, ATMs, traffic lights, household appliances, etc. are often discussed. Thus, bringing the advantages of \acp{SNN} to these sectors that relevant and present in the society would be another important advance.

Some possible improvements could be made over the proposed components. One of them would be the introduction of additional AND gates and signals to perform read operations in the memory. Currently, as explained in Section~\ref{memory_design_section}, the memory block design only allows write operations. Output spikes are held in its D latches, which are not currently expected to be connected to other neurons. In a practical application, it will be necessary to connect these latches to make proper use of the output spikes.

Furthermore, it could be interesting to reduce the amount of resources required to build some of the blocks presented in this work. Other improvement would be the elimination of the non-operation cases in the decoder, the encoder, the multiplexer and the demultiplexer. Note that non-operation cases mean that, while there are no input spikes, output spikes are generated or vice versa. The truth tables of these four components contemplate non-operation cases, which could be eliminated in order to remove their associated neurons and connections.

In \cite{ayuso2022spike}, the Constant Spike Source block is presented along with the energy efficiency problem. This problem could be an important subject of study, although it has been shown that there is evidence of a similar mechanism in biology \cite{douglas2007recurrent}.

One detail that has been taken into account when carrying out the implementation of the proposed blocks is that, since SpiNNaker only allows working with integer timesteps and ideal synapses, there should not be any unexpected deviation in the arrival time of input spikes. Thus, the problem of matching input spikes to perform certain operations, which is one of the weak points of asynchronous circuits, is inherently solved. This problem could be especially relevant for spiking AND gates, in which the precise timing of the spikes is necessary to produce the correct output. 

Notice that outside the SpiNNaker platform, this spike synchronization problem should be studied in depth, since it is intended for the designs to be as close as possible to the behavior of spikes and neurons in the nervous system. Said synchronization could be achieved in two different ways: the use of spike trains, which would surely imply changes in the presented designs, or the adjustment of the parameters used for the neurons to diminish the effect of small temporary variations in the arrival times of the spikes. This second option seems to be the most reasonable since it would involve few changes and could help to focus the synchronization problem in a low-level approach.

Finally, the spiking memory implementation presented in this work could have a large number of direct applications as its digital counterpart is essential in any computer, as explained in Section~\ref{intro_memory}. In fact, it allows to start thinking about the implementation of a spiking computer, which would be a great advance since it would be one of the highest-level spiking blocks implemented so far and would have the advantages of spiking blocks, which mainly are low power consumption and real-time capability.

\section{Conclusions}
\label{conclusions}

In this work, the spiking implementation of different digital components including decoder, encoder, multiplexer, demultiplexer, D Latch and memory is performed. The behavior of these blocks is proved by performing different tests, the results of which are shown to validate their performance.

In addition, the resources required to build each block presented in this paper are studied, showing different mathematical formulas that allow to quickly and easily obtain the number of neurons and synapses used in the implementation process and that can be very useful for neuromorphic engineers.

A deep comparison between the \ac{SNN}-based functional blocks proposed here and other alternatives in the state of the art have also been carried out, highlighting the main advantages and disadvantages and proving that it is a valid approach to design spiking functional blocks.

The blocks presented in this work are available in \mbox{sPyBlocks}, which is a public repository that could be very useful for neuromorphic engineers when completing neuromorphic applications, which may require certain complex functionalities and for which there was no \ac{SNN}-based solution.

On the other hand, some important points that are still open and the future work to be done using the tools provided are also studied.

The spiking memory implementation presented in this work paves the way in the development of spiking \acp{FSM} and a fully spiking computer.

\section*{Acknowledgments}
This research was partially supported by the Spanish grant MINDROB (PID2019-105556GB-C33/AEI/10.13039/501100011033). Daniel Casanueva-Morato was supported by a "Formación de Profesorado Universitario" Scholarship from the Spanish Ministry of Education, Culture and Sport.

\bibliographystyle{elsarticle-num} 
\bibliography{bibliography.bib}
 
\end{document}